\documentclass[10pt,british,10pt,twocolumn,letterpaper]{article}
\usepackage[T1]{fontenc}
\usepackage[latin9]{inputenc}
\usepackage{array}
\usepackage{rotating}
\usepackage{url}
\usepackage{enumitem}
\usepackage{multirow}
\usepackage{amsmath}
\usepackage{graphicx}

\makeatletter

\providecommand{\tabularnewline}{\\}
\newcommand{\lyxdot}{.}

\usepackage{cvpr}
\usepackage{times}
\usepackage{epsfig}
\usepackage{graphicx}
\usepackage{amsmath}
\usepackage{amssymb}

\usepackage{placeins}
\usepackage{hyphenat}
\usepackage{xspace}
\newcommand{\NMS}{NMS\xspace}
\newcommand{\gnet}{Gnet\xspace}

\usepackage{color}
\usepackage{colortbl}
\usepackage{rotating}
\definecolor{lightgray}{gray}{0.8}
\definecolor{verylightgray}{gray}{0.9}

\usepackage[pagebackref=true,breaklinks=true,letterpaper=true,colorlinks,bookmarks=false]{hyperref}

\cvprfinalcopy 

\def\cvprPaperID{1840} 

\ifcvprfinal\fi

\@ifundefined{showcaptionsetup}{}{%
 \PassOptionsToPackage{caption=false}{subfig}}
\usepackage{subfig}
\makeatother

\usepackage{babel}
\begin{document}
\makeatletter
\renewcommand{\paragraph}{%
\@startsection{paragraph}{4}%
 {\z@}{0.5ex \@plus 1ex \@minus .2ex}{-0.5em}%
  {\normalfont \normalsize \bfseries}%
}
\makeatother

\setlength{\textfloatsep}{1.5em}

\let\originalparagraph\paragraph
\renewcommand{\paragraph}[2][.]{\vspace{0.0em}\originalparagraph{#2#1}}

\title{Learning non-maximum suppression}

\author{Jan Hosang \and Rodrigo Benenson\vspace{0.5em}
\\
\begin{tabular}{c}
Max Planck Institut für Informatik\tabularnewline
Saarbrücken, Germany\tabularnewline
\texttt{\small{}firstname.lastname@mpi-inf.mpg.de}\tabularnewline
\end{tabular}\vspace{-0.5em}
 \and Bernt Schiele\\
}
\maketitle
\begin{abstract}
Object detectors have hugely profited from moving towards an end-to-end
learning paradigm: proposals, features, and the classifier becoming
one neural network improved results two-fold on general object detection.
One indispensable component is non-maximum suppression (\NMS), a
post-processing algorithm responsible for merging all detections that
belong to the same object. The de facto standard \NMS algorithm is
still fully hand-crafted, suspiciously simple, and \textemdash{} being
based on greedy clustering with a fixed distance threshold \textemdash{}
forces a trade-off between recall and precision. We propose a new
network architecture designed to perform \NMS, using only boxes and
their score. We report experiments for person detection on PETS and
for general object categories on the COCO dataset. Our approach shows
promise providing improved localization and occlusion handling.
\end{abstract}

\section{\label{sec:Introduction}Introduction}

All modern object detectors follow a three step recipe: (1) proposing
a search space of windows (exhaustive by sliding window or sparser
using proposals), (2) scoring/refining the window with a classifier/regressor,
and (3) merging windows that might belong to the same object. This
last stage is commonly referred to as ``non-maximum suppression''
(\NMS) \cite{Girshick2014Cvpr,Girshick2015IccvFastRCNN,Ren2015NipsFasterRCNN,Felzenszwalb2010Pami,Redmond2016Cvpr,Liu2016Eccv}.

The de facto standard for \NMS is a simple hand-crafted test time
post-processing, which we call GreedyNMS. The algorithm greedily selects
high scoring detections and deletes close-by less confident neighbours
since they are likely to cover the same object. This algorithm is
simple, fast, and surprisingly competitive compared to proposed alternatives.

\begin{figure}
\hfill{}\includegraphics[width=0.7\columnwidth]{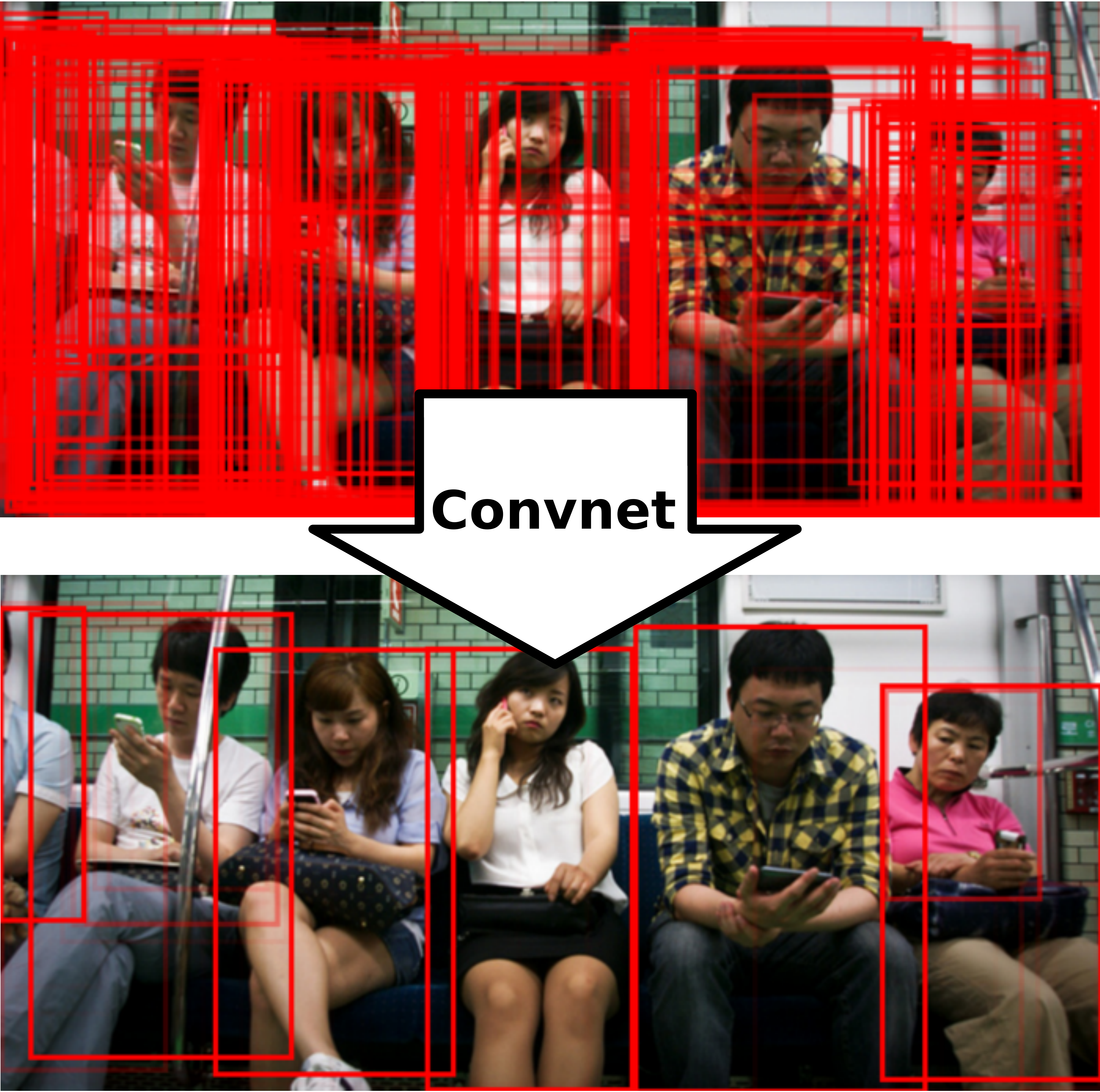}\hfill{}

\caption{We propose a non-maximum suppression convnet that will re-score all
raw detections (top). Our network is trained end-to-end to learn to
generate exactly one high scoring detection per object (bottom, example
result).}
\end{figure}

The most notable recent performance breakthrough in general object
detection was marked by R-CNN \cite{Girshick2014Cvpr}, which effectively
replaced features extraction and classifiers by a neural network,
almost doubling performance on Pascal VOC. Another significant improvement
was to absorb the object proposal generation into the network \cite{Ren2015NipsFasterRCNN},
while other works avoid proposals altogether \cite{Ren2015NipsFasterRCNN,Redmond2016Cvpr},
leading to both speed and quality improvements. We can see a general
trend towards end-to-end learning and it seems reasonable to expect
further improvements by doing complete end-to-end training of detectors.
\NMS is one step in the pipeline that, for the most part, has evaded
the end-to-end learning paradigm. All of the above detectors train
the classifier in a procedure that ignores the fact that the \NMS
problem exists and then runs GreedyNMS as a disconnected post-processing.

There is a need to overcome GreedyNMS due to its significant conceptual
shortcomings. GreedyNMS makes hard decision by deleting detections
and bases this decision on one fixed parameter that controls how wide
the suppression is. A wide suppression would remove close-by high
scoring detections that are likely to be false positives that hurt
precision. On the other hand, if objects are close (e.g. in crowded
scenes), close-by detections can be true positives, in which case
suppression should be narrow to improve recall. When objects are close-by,
GreedyNMS is doomed to sacrifice precision or recall independent of
its parameter.

It is desirable to learn \NMS to overcome these limitations. An NMS
approach based on neural network could learn to adapt to the data
distribution, overcome the trade-off of GreedyNMS, and importantly
could be incorporated \emph{into} a detector. In this paper we propose
the first ``pure \NMS network'' which is able to do the task of
non-maximum suppression without image content or access to decisions
of another algorithm. Our network renders the need for final GreedyNMS
post-processing superfluous.

In section \ref{sec:NMS} we start by discussing with the underlying
issue: why is \NMS needed at all? We discuss the task of detection
and how it relates to the specifics of detectors and \NMS. We identify
two necessary ingredients that current detectors are lacking and design
an \NMS network that contains these ingredients (section \ref{sec:Approach});
the result is conceptually different than both GreedyNMS and current
detectors. In section \ref{sec:experiments}, we report promising
results that show that this network is indeed capable of replacing
GreedyNMS. We report both single- (PETS pedestrians) and multi-class
results (COCO dataset), both showing improvements over GreedyNMS.

We believe this work opens the door to true end-to-end detectors.

\section{\label{subsec:Related-work}Related work}

\paragraph{Clustering detections}

The de facto standard algorithm, GreedyNMS, has survived several generations
of detectors, from Viola\&Jones \cite{Viola2004Ijvc}, over the deformable
parts model (DPM) \cite{Felzenszwalb2010Pami}, to the current state-of-the-art
R-CNN family \cite{Girshick2014Cvpr,Girshick2015IccvFastRCNN,Ren2015NipsFasterRCNN}.
Several other clustering algorithms have been explored for the task
of \NMS without showing consistent gains: mean-shift clustering \cite{Dalal2005Cvpr,Wojek2008Pr},
agglomerative clustering \cite{Bourdev2010Eccv}, affinity propagation
clustering \cite{mrowca2015Iccv}, and heuristic variants \cite{Sermanet2014Iclr}.
Principled clustering formulations with globally optimal solutions
have been proposed in \cite{Tang2015Cvpr,Rothe2015Accv}, although
they have yet to surpass the performance of GreedyNMS.

\paragraph{Linking detections to pixels}

Hough voting establishes correspondences between detections and the
image evidence supporting them, which can avoid overusing image content
for several detections \cite{Leibe2008Ijcv,Barinova2012Pami,Kontschieder2012CVIU,Wohlhart2012Accv}.
Overall performance of hough voting detectors remains comparatively
low. \cite{Yao2012Cvpr,Dai2015Cvpr} combine detections with semantic
labelling, while \cite{Yan2015Cvpr} rephrase detection as a labelling
problem. Explaining detections in terms of image content is a sound
formulation but these works rely on image segmentation and labelling,
while our system operates purely on detections without additional
sources of information.

\paragraph{Co-occurrence}

One line of work proposes to detect pairs of objects instead of each
individual objects in order to handle strong occlusion \cite{Sadeghi2011Cvpr,Tang2012Bmvc,Ouyang2013Cvpr}.
It faces an even more complex \NMS problem, since single and double
detections need to be handled. \cite{Rodriguez2011Iccv} bases suppression
decisions on estimated crowd density. Our method does neither use
image information nor is it hand-crafted to specifically detect pairs
of objects.

\paragraph{Auto-context}

Some methods improve object detection by jointly rescoring detections
locally \cite{Tu2010Pami,Chen2013CvprMoco} or globally \cite{Vezhnevets2015Bmvc}
using image information. These approaches tend to produce fewer spread-out
double detections and improve overall detection quality, but still
require \NMS. We also approach the problem of \NMS as a rescoring
task, but we completely eliminate any post-processing.

\paragraph{Neural networks on graphs}

A set of detections can be seen as a graph where overlapping windows
are represented as edges in a graph of detections. \cite{niepert2016icml}
operates on graphs, but requires a pre-processing that defines a node
ordering, which is ill-defined in our case.

\paragraph{End-to-end learning for detectors}

Few works have explored true end-to-end learning that includes \NMS.
One idea is to include GreedyNMS at training time \cite{Wan2015Cvpr,Henderson2016ACCV},
making the classifier aware of the NMS procedure at test time. This
is conceptually more satisfying, but does not make the \NMS learnable.
Another interesting idea is to directly generate a sparse set of detections,
so \NMS is unnecessary, which is done in \cite{Stewart2015Arxiv}
by training an LSTM that generates detections on overlapping patches
of the image. At the boundaries of neighbouring patches, objects might
be predicted from both patches, so post-processing is still required.
\cite{Hosang2016Gcpr} design a convnet that combines decisions of
GreedyNMS with different overlap thresholds, allowing the network
to choose the GreedyNMS operating point locally. None of these works
actually completely \emph{remove} GreedyNMS from the final decision
process that outputs a sparse set of detections. Our network is capable
of performing \NMS without being given a set of suppression alternatives
to chose from and without having another final suppression step.

\section{\label{sec:NMS}Detection and non-maximum suppression}

In this section we review non-maximum suppression (\NMS) and why
it is necessary. In particular, we point out why current detectors
are conceptually incapable of producing exactly one detection per
object and propose two necessary ingredients for a detector to do
so.

Present-day detectors do not return all detections that have been
scored, but instead use \NMS as a post-processing step to remove
redundant detections. In order to have true end-to-end learned detectors,
we are interested in detectors without any post-processing. To understand
why \NMS is necessary, it is useful to look at the task of detection
and how it is evaluated.

\paragraph{Object detection}

The task of object detection is to map an image to a set of boxes:
one box per object of interest in the image, each box tightly enclosing
an object. This means detectors ought to return exactly one detection
per object. Since uncertainty is an inherent part of the detection
process, evaluations allow detections to be associated to a confidence.
Confident erroneous detections are penalized more than less confident
ones. In particular mistakes that are less confident than the least
confident correct detection are not penalized at all.

\paragraph{Detectors do not output what we want}

The detection problem can be interpreted as a classification problem
that estimates probabilities of object classes being present for every
possible detection in an image. This viewpoint gives rise to ``hypothesize
and score'' detectors that build a search space of detections (e.g.~sliding
window, proposals) and estimate class probabilities \emph{independently}
for each detection. As a result, two strongly overlapping windows
covering the same object will both result in high score since they
look at almost identical image content. In general, instead of one
detection per object, each object triggers several detections of varying
confidence, depending on how well the detection windows cover the
object.

\paragraph{GreedyNMS}

Since the actual goal is to generate \emph{exactly one} detection
per object (or exactly one high confidence detection), a common practice
(since at least 1994 \cite{Burel1994Prl}) is to assume that highly
overlapping detections belong to the same object and collapse them
into one detection. The predominant algorithm (GreedyNMS) accepts
the highest scoring detection, then rejects all detections that overlap
more than some threshold $\vartheta$ and repeats the procedure with
the remaining detections, i.e.~greedily accepting local maxima and
discarding their neighbours, hence the name. This algorithm eventually
also accepts wrong detections, which is no problem if their confidence
is lower than the confidence of correct detections.

\paragraph{GreedyNMS is not good enough}

This algorithm works well if (1) the suppression is wide enough to
always suppress high scoring detections triggered by same object and
(2) the suppression is narrow enough to never suppress high scoring
detections of the next closest object. If objects are far apart condition
(2) is easy to satisfy and a wide suppression works well. In crowded
scenes with high occlusion between objects there is a tension between
wide and narrow suppression. In other words with one object per image
\NMS is trivial, but highly occluded objects require a better \NMS
algorithm.

\subsection{A future without \NMS}

Striving for true end-to-end systems without hand crafted algorithms
we shall ask: \textbf{Why do we need a hand crafted post processing
step? Why does the detector not directly output one detection per
object?}

Independent processing of image windows leads to overlapping detection
giving similar scores, this is a requirement of robust functions:
similar inputs lead to similar outputs. A detector that outputs only
one high scoring detection per object thus has to be also conditioned
on other detections: multiple detections on the same object should
be processed jointly, so the detector can tell there are repeated
detections and only one of them should receive a high score.

Typical inference of detectors consist of a classifier that discriminates
between image content that contains an object and image content that
does not. The positive and negative training examples for this detector
are usually defined by some measure of overlap between objects and
bounding boxes. Since similar boxes will produce similar confidences
anyway, small perturbation of object locations can be considered positive
examples, too. This technique augments the training data and leads
to more robust detectors. Using this type of classifier training does
not reward one high scoring detection per object, and instead deliberately
encourages multiple high scoring detections per object.

From this analysis we can see that two key ingredients are necessary
in order for a detector to generate exactly one detection per object:
\begin{enumerate}
\item A \emph{loss} that penalises double detections to teach the detector
we want \emph{precisely one} detection per object.
\item \emph{Joint processing} of neighbouring detections so the detector
has the necessary information to tell whether an object was detected
multiple times.
\end{enumerate}
In this paper, we explore a network design that accommodates both
ingredients. To validate the claim that these are key ingredients
and our the proposed network is capable of performing NMS, we study
our network in isolation without end-to-end learning with the detector.
That means the network operates solely on scored detections without
image features and as such can be considered a ``pure NMS network''.

\section{\label{sec:Approach}Doing NMS with a convnet}

After establishing the two necessary requirements for a convnet (convolutional
network) to perform NMS in section \ref{sec:NMS}, this section presents
our network that addresses both (penalizing double detections in \S\ref{subsec:Loss},
joint processing of detections in \S\ref{subsec:Chatty-windows}).

Our design avoids hard decisions and does not discard detections to
produce a smaller set of detections. Instead, we reformulate NMS as
a rescoring task that seeks to decrease the score of detections that
cover objects that already have been detected, as in \cite{Hosang2016Gcpr}.
After rescoring, simple thresholding is sufficient to reduce the set
of detections. For evaluation we pass the full set of rescored detections
to the evaluation script without any post processing.

\subsection{\label{subsec:Loss}Loss}

A detector is supposed to output exactly one high scoring detection
per object. The loss for such a detector must inhibit multiple detections
of the same object, irrespective of how close these detections are.
Stewart and Andriluka \cite{Stewart2015Arxiv} use a Hungarian matching
loss to accomplish that: successfully matched detections are positives
and unmatched detections are negatives. The matching ensures that
each object can only be detected once and any further detection counts
as a mistake. Henderson and Ferrari \cite{Henderson2016ACCV} present
an average precision (AP) loss that is also based on matching.

Ultimately a detector is judged by the evaluation criterion of a benchmark,
which in turn defines a matching strategy to decide which detections
are correct or wrong. This is the matching that should be used at
training time. Typically benchmarks sort detections in descending
order by their confidence and match detections in this order to objects,
preferring most overlapping objects. Since already matched objects
cannot be matched again surplus detections are counted as false positives
that decrease the precision of the detector. We use this matching
strategy.

We use the result of the matching as labels for the classifier: successfully
matched detections are positive training examples, while unmatched
detections are negative training examples for a standard binary loss.
Typically all detections that are used for training of a classifier
have a label associated as they are fed into the network. In this
case the network has access to detections and object annotations and
the matching layer generates labels, that depend on the predictions
of the network. Note how this class assignment directly encourages
the rescoring behaviour that we wish to achieve.

Let $d_{i}$ denote a detection, $y_{i}\in\{-1,1\}$ indicate whether
or not $d_{i}$ was successfully matched to an object, and let $f$
denote the scoring function that jointly scores all detections on
an image $f\left(\left[d_{i}\right]_{i=1}^{n}\right)=\left[s_{i}\right]_{i=1}^{n}$.
We train with the weighted logistic loss
\[
L(s_{i},y_{i})=\sum_{i=1}^{N}w_{y_{i}}\cdot\log\left(1+\exp\left(-s_{i}\cdot y_{i}\right)\right).
\]

Here loss per detection is coupled to the other detections through
the matching that produces $y_{i}$.

The weighting $w_{y_{i}}$ is used to counteract the extreme class
imbalance of the detection task. We choose the weights so the expected
class conditional weight of an example equals a parameter $\mathbf{E}\left(w_{1}I\left(y_{i}=1\right)\right)=\gamma$.

When generalising to the \textbf{multiclass} setting, detections are
associated to both a confidence and a class. Since we only rescore
detections, we allow detections to be ``switched off'' but not to
change their class. As a result, we only match detections to objects
of the same class, but the classification problem remains binary and
the above loss still applies. When representing the detection scores,
we use a one-hot encoding: a zero vector that only contains the score
at the location in the vector that corresponds to the class. Since
mAP computation does not weight classes by their size, we assign the
instance weights in a way that their expected class conditional weight
is uniformly distributed.

\subsection{\label{subsec:Chatty-windows}``Chatty'' windows}

\begin{figure*}
\centering{}\includegraphics[width=1\textwidth]{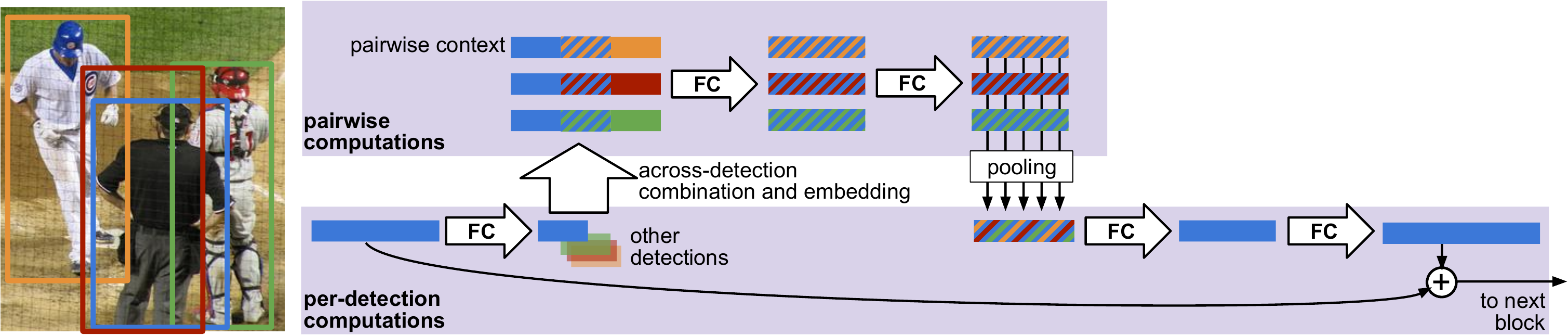}\caption{\label{fig:network-architecture}One block of our \gnet visualised
for \emph{one} detection. The representation of each detection is
reduced and then combined into neighbouring detection pairs and concatenated
with detection pair features (hatched boxes, corresponding features
and detections have the same colour). Features of detection pairs
are mapped independently through fully connected layers. The variable
number of pairs is reduced to a fixed-size representation by max-pooling.
Pairwise computations are done for each detection independently.}
\end{figure*}

In order to effectively minimize the aforementioned loss, we need
our network to jointly process detections. To this end we design a
network with a repeating structure, which we call blocks (sketched
in figure \ref{fig:network-architecture}). One block gives each detection
access to the representation of its neighbours and subsequently updates
its own representation. Stacking multiple blocks means the network
alternates between allowing every detection ``talk'' to its neighbours
and updating its own representation. We call this the \textbf{GossipNet}
(\gnet), because detections talk to their neighbours to update their
representation.

There are two non-standard operations here that are key. The first
is a layer, that builds representations for pairs of detections. This
leads to the key problem: an irregular number of neighbours for each
detection. Since we want to avoid the discretisation scheme used in
\cite{Hosang2016Gcpr}, we will solve this issue with pooling across
detections (the second key).

\begin{figure}
\hfill{}\subfloat[\label{fig:gnet-high-level-train}Training architecture.]{\begin{centering}
\includegraphics[width=1\columnwidth]{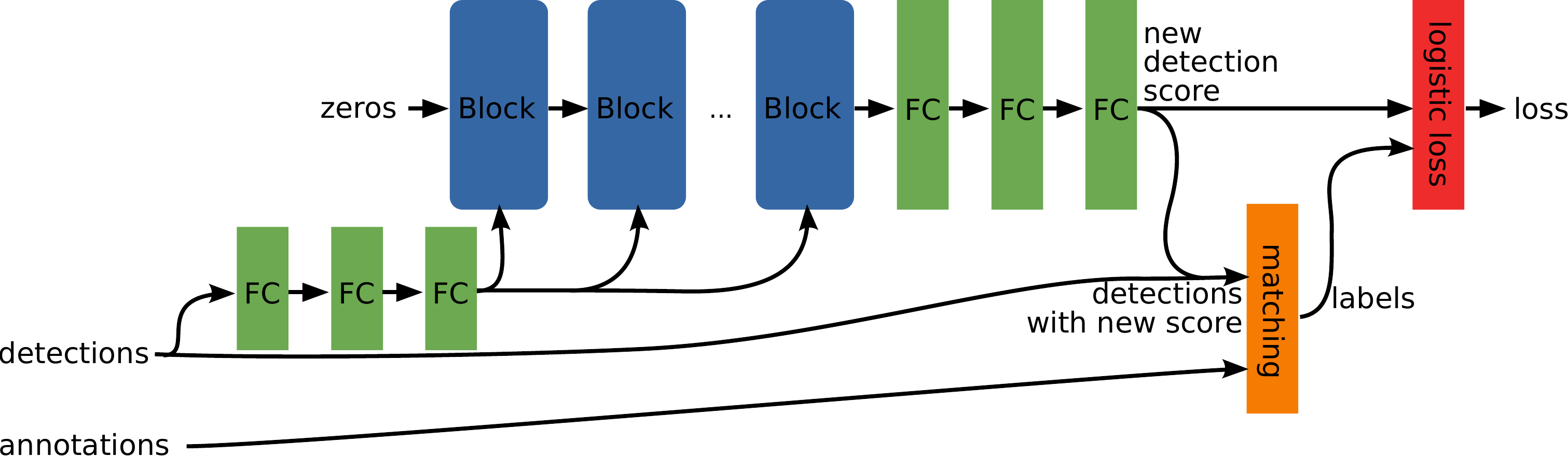}
\par\end{centering}
\centering{}}\hfill{}

\hfill{}\subfloat[\label{fig:gnet-high-level-test}Test architecture.]{\begin{centering}
\includegraphics[width=0.8\columnwidth]{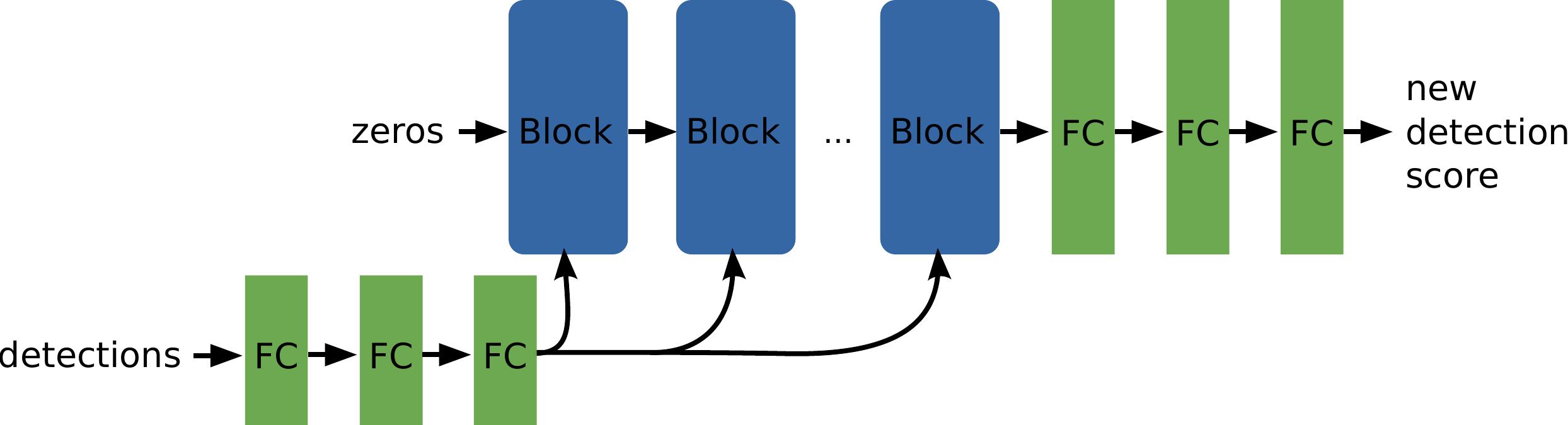}
\par\end{centering}
}\hfill{}

\caption{\label{fig:gnet-high-level}High level diagram of the Gnet. FC denotes
fully connected layers. All features in this diagram have 128 dimensions
(input vector and features between the layers/blocks), the output
is a scalar.}
\end{figure}

\paragraph{Detection features}

The blocks of our network take the detection feature vector of each
detection as input and outputs an updated vector (see high-level illustration
in figure~\ref{fig:gnet-high-level}). Outputs from one block are
input to the next one. The values inside this $c=128$ dimensional
feature vector are learned implicitly during the training. The output
of the last block is used to generate the new detection score for
each detection.

The first block takes an all-zero vector as input. The detections'
information is fed into the network in the ``pairwise computations''
section of figure~\ref{fig:network-architecture} as described below.
In future work this zero input could potentially be replaced with
image features.

\paragraph{Pairwise detection context}

Each mini-batch consists of all $n$ detections on an image, each
represented by a $c$ dimensional feature vector, so the data has
size $n\times c$ and accessing to another detection's representations
means operating within the batch elements. We use a detection context
layer, that, for every detection $d_{i}$, generates all pairs of
detections $\left(d_{i},d_{j}\right)$ for which $d_{j}$ sufficiently
overlaps with $d_{i}$ (IoU > 0.2). The representation of a pair of
detections consists of the concatenation of both detection representations
and $g$ dimensional detection pair features (see below), which yields
an $l=2c+g$ dimensional feature. To process each pair of detections
independently, we arrange the features of all pairs of detections
along the batch dimension: if detection $d_{i}$ has $k_{i}$ neighbouring
detection that yields a batch of size $K\times l$, where $K=\sum_{i=1}^{n}(k_{i}+1)$
since we also include the pair $(d_{i},d_{i})$. Note that the number
of neighbours $k_{i}$ (the number of pairs) is different for every
detection even within one mini-batch. To reduce the variable sized
neighbourhood into a fixed size representation, our architecture uses
global max-pooling over all detection pairs that belong to the same
detection ($K\times l\rightarrow n\times l$), after which we can
use normal fully connected layers to update the detection representation
(see figure~\ref{fig:network-architecture}).

\paragraph{Detection pair features}

The features for each detection pair used in the detection context
consists of several properties of a detection pair: (1) the intersection
over union (IoU), (2-4) the normalised distance in x and y direction
and the normalised l2 distance (normalized by the average of width
and height of the detection), (4-5) scale difference of width and
height (e.g.~$\log\left(w_{i}/w_{j}\right)$), (6) aspect ratio difference
$\log\left(a_{i}/a_{j}\right)$, (7-8) the detection scores of both
detections. In the multi-class setup, each detection provides a scores
vector instead of a scalar thus increasing the number of pair features.
We feed all these raw features into 3 fully connected layers, to learn
the $g$ detection pair features that are used in each block.

\paragraph{Block}

A block does one iteration allowing detections to look at their respective
neighbours and updating their representation (sketched in figure \ref{fig:network-architecture}).
It consists of a dimensionality reduction, a pairwise detection context
layer, 2 fully connected layers applied to each pair independently,
pooling across detections, and two fully connected layers, where the
last one increases dimensionality again. The input and output of a
block are added as in the Resnet architecture \cite{He2016Eccv}.
The first block receives zero features as inputs, so all information
that is used to make the decision is bootstrapped from the detection
pair features. The output of the last block is used by three fully
connected layers to predict a new score for each detection independently
(figure~\ref{fig:gnet-high-level}).

\paragraph{Parameters}

Unless specified otherwise our networks have 16 blocks. The feature
dimension for the detection features is 128 and is reduced to 32 before
building the pairwise detection context. The detection pair features
also have 32 dimensions. The fully connected layers after the last
block output 128 dimensional features. When we change the feature
dimension, we keep constant the ratio between the number of features
in each layer, so indicating the detection feature dimension is sufficient.

\paragraph{Message passing}

The forward pass over serveral stacked blocks can be interpreted as
message passing. Every detection sends messages to all of its neighbours
in order to negotiate which detection is assigned an object and which
detections should decrease their scores. Instead of hand-crafting
the message passing algorithm and its rules, we deliberately let the
network latently learn the messages that are being passed.

\subsection{Remarks}

The \gnet is fundamentally different than GreedyNMS in the sense
that all features are updated concurrently, while GreedyNMS operates
sequentially. Since \gnet does not have access to GreedyNMS decisions
(unlike \cite{Hosang2016Gcpr}), it is surprising how close performance
of the two algorithms turns out to be in section \ref{sec:experiments}.
Since we build a potentially big network by stacking many blocks,
the \gnet might require large amounts of training data. In the experiments
we deliberately choose a setting with many training examples.

The \gnet is a pure NMS network in the sense that it has no access
to image features and operates solely on detections (box coordinates
and a confidence). This means the \gnet cannot be interpreted as
extra layers to the detector. The fact that it is a neural network
and that it is possible to feed in a feature vector (from the image
or the detector) into the first block makes it particularly suitable
for combining it with a detector, which we leave for future work.

The goal is to \emph{jointly} rescore all detections on an image.
By allowing detections to look at their neighbours and update their
own representation, we enable conditional dependence between detections.
Together with the loss that encourages exactly one detection per object,
we have satisfied both conditions from section \ref{sec:NMS}. We
will see in section \ref{sec:experiments} that the performance is
relatively robust to parameter changes and works increasingly well
for increasing depth.

\section{\label{sec:experiments}Experiments}

In this section we experimentally evaluate the proposed architecture
on the PETS and COCO dataset. We report results for persons, and as
well for the multi-class case. Person category is by far the largest
class on COCO, and provides both crowded images and images with single
persons. Other than overall results, we also report separately high
and low occlusion cases. We are interested in performance under occlusion,
since this is the case in which non-maximum suppression (\NMS) is
hard. All-in-all we show a consistent improvement over of GreedyNMS,
confirming the potential of our approach.

All results are measured in average precision (AP), which is the area
under the recall-precision curve. The overlap criterion (for matching
detections to objects) is traditionally 0.5 IoU (as for Pascal VOC,
noted as $\text{AP}{}_{0.5}$). COCO also uses stricter criteria to
encourage better localisation quality, one such metric averages AP
evaluated over several overlap criteria in the range $[0.5,0.95]$
in $0.05$ increments, which we denote by $\text{AP}{}_{0.5}^{0.95}$.

\subsection{\label{subsec:PETS-results}PETS: Pedestrian detection in crowds}

\begin{figure}
\centering{}\includegraphics[width=0.9\columnwidth]{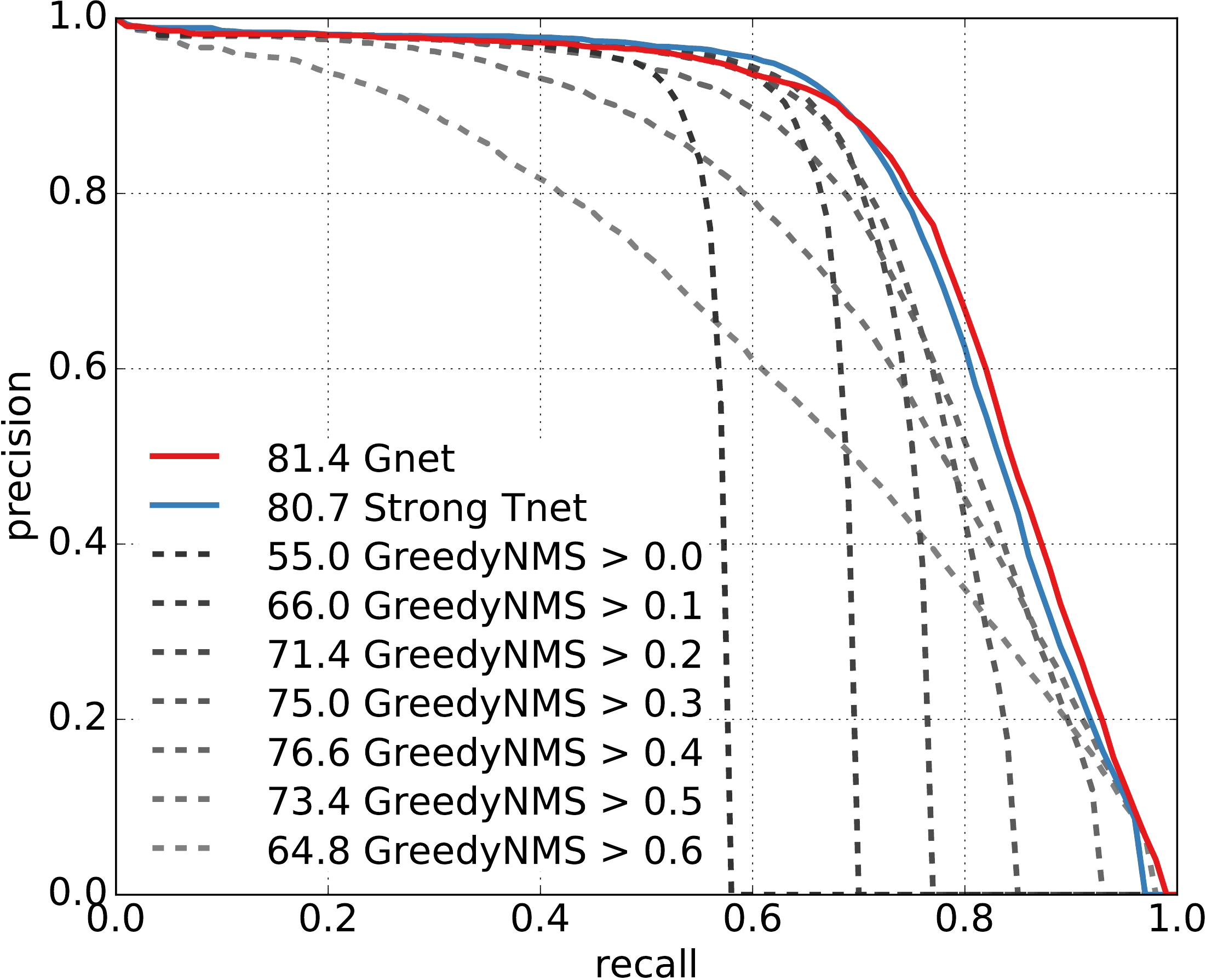}\caption{\label{fig:Pets-results}Performance on the PETS test set.}
\end{figure}
\begin{figure}
\centering{}\includegraphics[width=1\columnwidth]{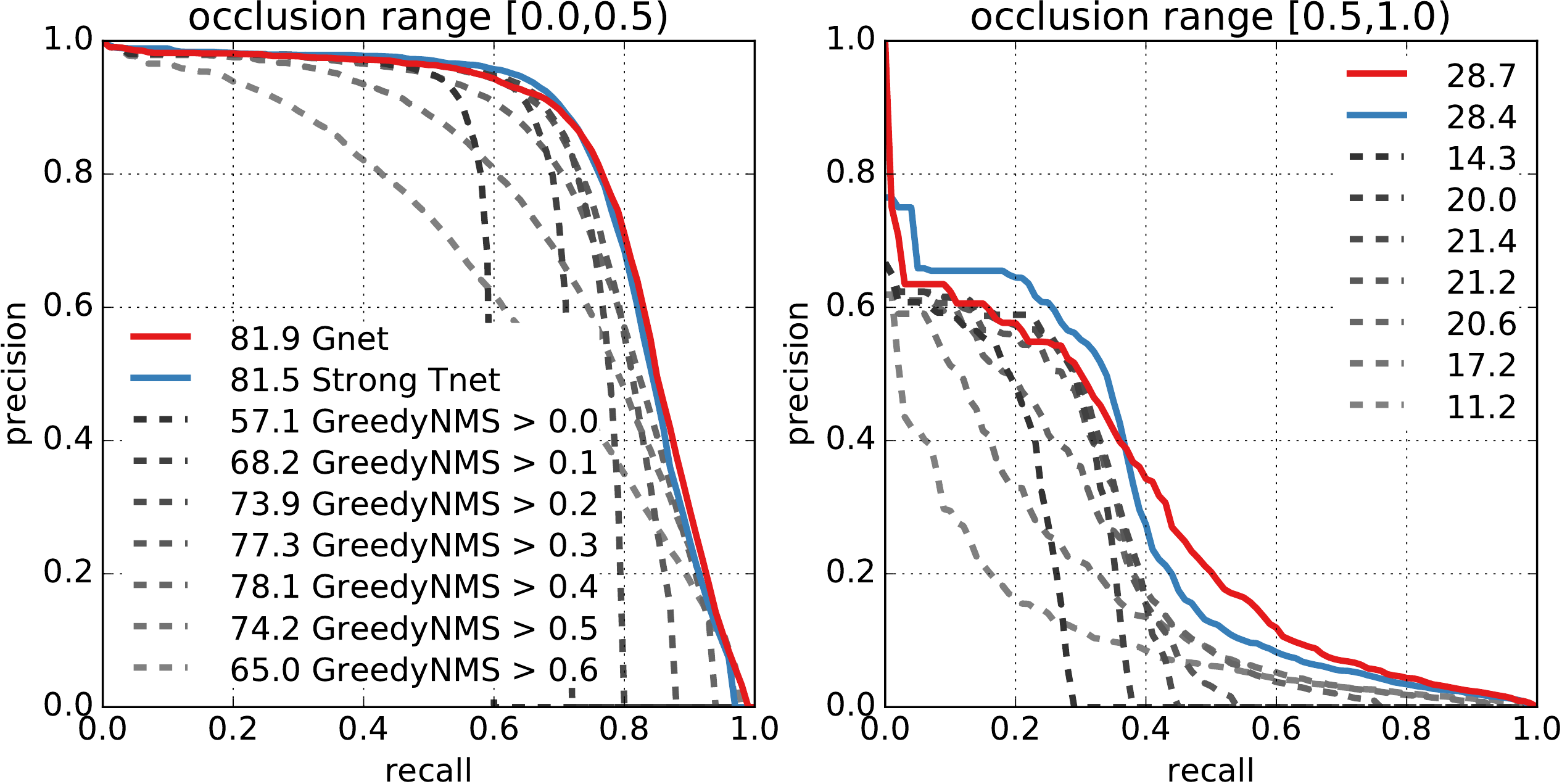}\caption{\label{fig:Pets-occlusion}Performance on the PETS test set for different
occlusion ranges.}
\end{figure}

\paragraph{Dataset}

PETS \cite{Ferryman2010AvssPetsDataset} is a dataset consisting of
several crowded sequences. It was used in \cite{Hosang2016Gcpr} as
a roughly single scale pedestrian detection dataset with diverse levels
of occlusion. Even though we aim for a larger and more challenging
dataset we first analyse our method in the setup proposed in \cite{Hosang2016Gcpr}.
We use the same training and test set as well as the same detections
from \cite{Tang2013Iccv}, a model built specifically to handle occlusions.
We reduce the number of detections with an initial GreedyNMS of 0.8
so we can fit the joint rescoring of all detections into one GPU.
(Note that these detections alone lead to bad results, worse than
``GreedyNMS > 0.6'' in \ref{fig:Pets-results}, and this is very
different to having input of GreedyNMS of 0.5 as an input like in
\cite{Hosang2016Gcpr}).

\paragraph{Training}

We train a model with 8 blocks and a 128 dimensional detection representation
for 30k iterations, starting with a learning rate of $10^{-3}$ and
decrease it by $0.1$ every 10k iterations.

\paragraph{Baselines}

We compare to the (typically used) classic GreedyNMS algorithm using
several different overlap thresholds, and the Strong Tnet from \cite{Hosang2016Gcpr}.
Since all methods operate on the same detections, the results are
fully comparable.

\paragraph{Analysis}

Figure \ref{fig:Pets-results} compares our method with the GreedyNMS
baseline and the Tnet on the PETS test set. Starting from a wide GreedyNMS
suppression with the threshold $\vartheta=0$ shows almost a step
function, since high scoring true positives suppress all touching
detections at the cost of also suppressing other true positives (low
recall). Gradually increasing $\vartheta$ improves the maximum recall
but also introduces more high scoring false positives, so precision
is decreasing. This shows nicely the unavoidable trade-off due to
having a fixed threshold $\vartheta$ mentioned in section \ref{sec:NMS}.
The reason for the clear trade-off is the diverse occlusion statistics
present in PETS.

Tnet performs better than the upper envelope of the GreedyNMS, as
it essentially recombines output of GreedyNMS at a range of different
thresholds. In comparison our \gnet performs slightly better, despite
not having access to GreedyNMS decisions at all. Compared to the best
GreedyNMS performance, \gnet is able to improve by 4.8 AP.

Figure \ref{fig:Pets-occlusion} shows performance separated into
high and low occlusion cases. Again, the \gnet performs slightly
better than Tnet. Performance in the occlusion range $[0,0.5)$ looks
very similar to the performance overall. For the highly occluded cases,
the performance improvement of \gnet compared to the best GreedyNMS
is bigger with 7.3 AP. This shows that the improvement for both \gnet
and Tnet is mainly due to improvements on highly occluded cases as
argued in section \ref{sec:NMS}.

\subsection{\label{subsec:COCO-results}COCO: Person detection }

\paragraph{Dataset}

The COCO datasets consists of 80k training and 40k evaluation images.
It contains 80 different categories in unconstrained environments.
We first mimic the PETS setup and evaluate for persons only, and report
multi-class results in section \ref{subsec:COCO-multiclass}.

Since annotations on the COCO test set are not available and we want
to explicitly show statistics per occlusion level, we train our network
on the full training set and evaluate using two different subsets
of the validation set. One subset is used to explore architectural
choices for our network (minival, 5k images\footnote{We use the same as used by Ross Girshick \url{https://github.com/rbgirshick/py-faster-rcnn/tree/master/data}.})
and the most promising model is evaluated on the rest of the validation
set (minitest, 35k images).

We use the Python implementation of Faster R-CNN \cite{Ren2015NipsFasterRCNN}\footnote{\url{https://github.com/rbgirshick/py-faster-rcnn}}
for generating detections. We train a model only on the training set,
so performance is slightly different than the downloadable model,
which has been trained on the training and minitest sets. We run the
detector with default parameters, but lower the detection score threshold
and use detection before the typical non-maximum suppression step.
There is no further preprocessing.

\paragraph{Training}

We train the \gnet with ADAM for $2\cdot10^{6}$ iterations, starting
with a learning rate of $10^{-4}$ and decreasing it to $10^{-5}$
after $10^{6}$ iterations. The detection feature dimension is 128,
the number of blocks is specified for each experiment.

\paragraph{Speed}

On average we have 67.3 person detection per image, which the 16 block
\gnet can process in 14ms/image on a K40m GPU and unoptimised Tensorflow
code.

\paragraph{Baselines}

We use GreedyNMS as a baseline. To show it in its best light we tune
the optimal GreedyNMS overlap threshold on the test set of each experiment.
\begin{figure}
\centering{}\includegraphics[width=1\columnwidth]{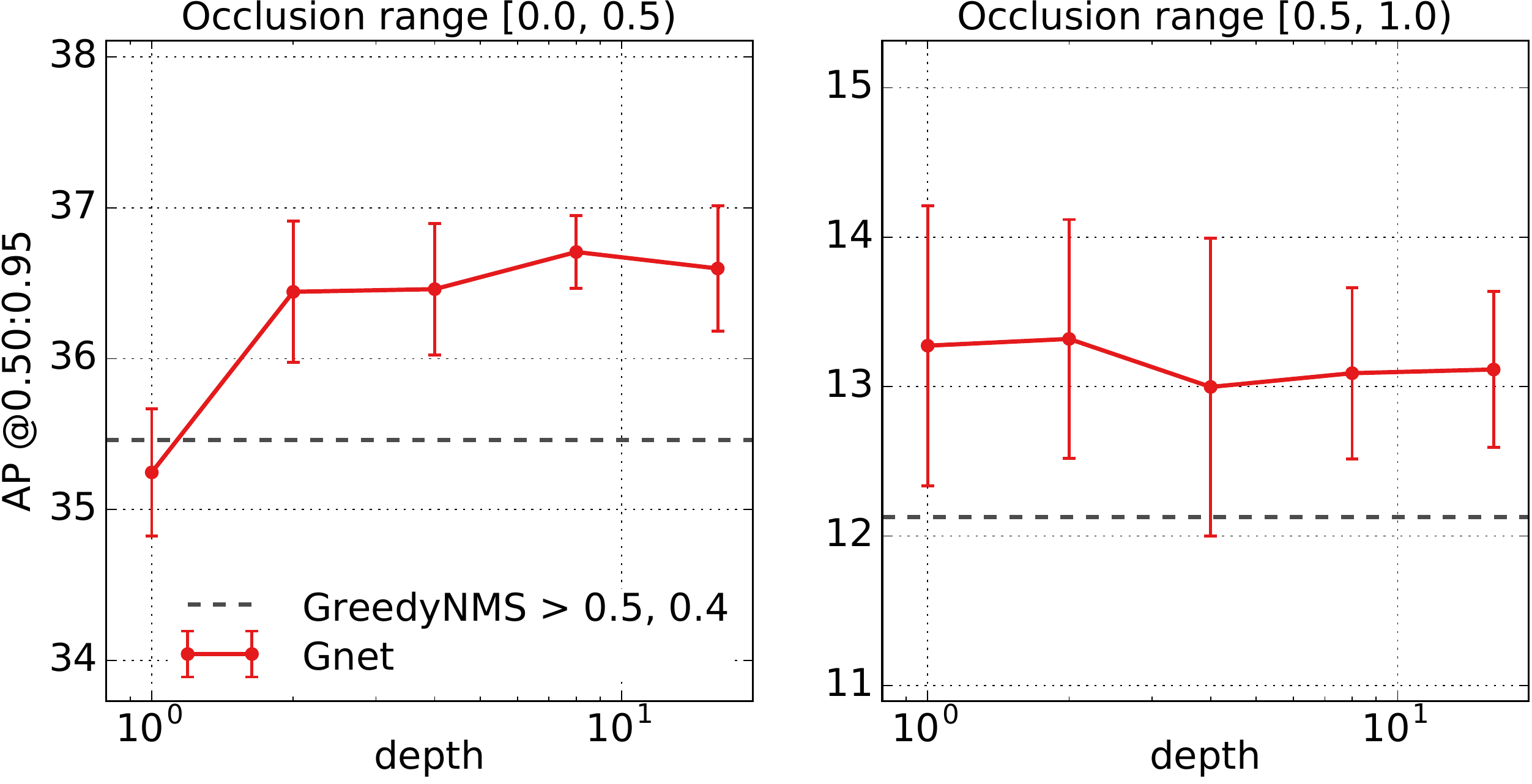}\caption{\label{fig:number-of-blocks-analysis}$\text{AP}{}_{0.5}^{0.95}$
versus number of blocks ($2,4,8,16$) for low and high occlusion respectively
on COCO persons minival. Average over six runs, error bars show the
standard deviation.}
\end{figure}
\begin{table}
\begin{centering}
\setlength{\tabcolsep}{0.1em}
\renewcommand{\arraystretch}{1}%
\begin{tabular}{cc|cc|cc|cc}
 &  & \multicolumn{2}{c|}{All} & \multicolumn{2}{c|}{%
\begin{tabular}{c}
Occlusion \tabularnewline
{[}0, 0.5)\tabularnewline
\end{tabular}} & \multicolumn{2}{c}{%
\begin{tabular}{c}
Occlusion \tabularnewline
{[}0.5, 1{]}\tabularnewline
\end{tabular}}\tabularnewline
 & Method & $\text{AP}{}_{0.5}$ & $\text{AP}{}_{0.5}^{0.95}$ & $\text{AP}{}_{0.5}$ & $\text{AP}{}_{0.5}^{0.95}$ & $\text{AP}{}_{0.5}$ & $\text{AP}{}_{0.5}^{0.95}$\tabularnewline
\hline
\hline
\multirow{2}{*}{\begin{turn}{90}
{\small{}val\hspace*{0.25em}}
\end{turn}} & {\small{}GreedyNMS>0.5} & 65.6 & 35.6 & 65.2 & 35.2 & 35.3 & 12.1\tabularnewline
 & {\small{}\gnet, 8 blocks} & \textbf{67.3} & \textbf{36.9} & \textbf{66.9} & \textbf{36.7} & \textbf{36.7} & \textbf{13.1}\tabularnewline
\hline
\multirow{2}{*}{\begin{turn}{90}
{\small{}test\hspace*{0.5em}}
\end{turn}} & {\small{}GreedyNMS>0.5} & 65.0 & 35.5 & 61.8 & 33.8 & 30.3 & 11.0\tabularnewline
 & {\small{}\gnet, 8 blocks} & \textbf{66.6} & \textbf{36.7} & \textbf{66.8} & \textbf{36.1} & \textbf{33.9} & \textbf{12.4}\tabularnewline
\end{tabular}
\par\end{centering}
\caption{\label{tab:COCO-persons-results}Comparison between \gnet and GreedyNMS
on COCO persons minival and minitest. Results for the full set and
separated into occlusion levels.}
\end{table}

\paragraph{Analysis}

Figure \ref{fig:number-of-blocks-analysis} shows $\text{AP}{}_{0.5}^{0.95}$
versus number of blocks in \gnet. The optimal GreedyNMS thresholds
are $0.5$ and $0.4$ for low and high occlusion respectively. Already
with one block our network performs on par with GreedyNMS, with two
blocks onwards we see a $\sim\negmedspace1\ \text{AP}$ point gain.
As in PETS we see gains both for low and high occlusions. With deeper
architectures the variance between models for the high occlusion case
seems to be decreasing, albeit we expect to eventually suffer from
over-fitting if the architecture has too many free parameters.

We conclude that our architecture is well suited to replace GreedyNMS
and is not particularly sensitive to the number of blocks used. Table
\ref{tab:COCO-persons-results} shows detailed results for \gnet
with 8 blocks. The results from the validation set (minival) transfer
well to the test case (minitest), providing a small but consistent
improvement over a well tuned GreedyNMS. Qualitative results are included
in the supplementary material.

We consider these encouraging results, confirming that indeed the
\gnet is capable of properly performing \NMS without access to image
features or GreedyNMS decisions.

\subsection{\label{subsec:COCO-multiclass}COCO multi-class}

\begin{figure}
\centering{}\includegraphics[width=1\columnwidth]{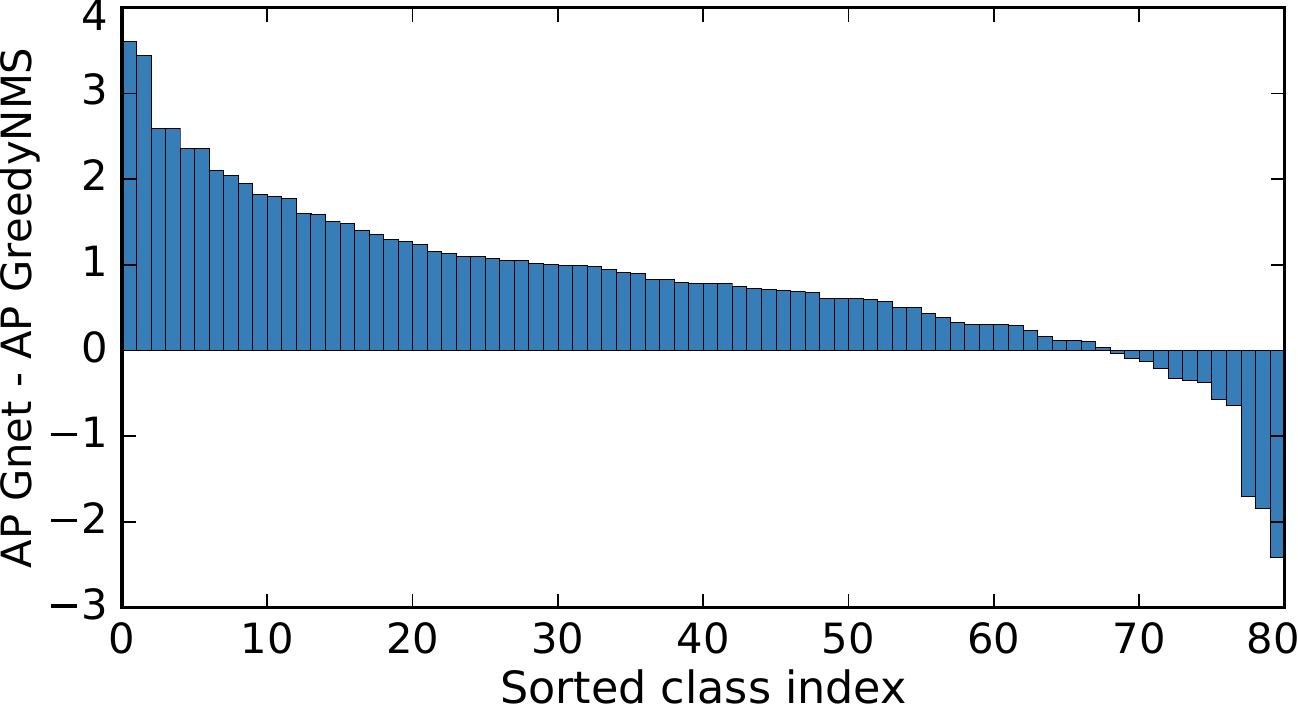}\caption{\label{fig:multiclass-minitest}$\text{AP}{}_{0.5}^{0.95}$ improvement
of \gnet over the best GreedyNMS threshold for each of the (sorted)
80 COCO classes. \gnet improves on $\sim\negmedspace70$ out of 80
categories. On average \gnet provides a $\sim\negmedspace1\ \text{mAP}{}_{0.5}^{0.95}$
point gain over per-class tuned GreedyNMS ($23.5\rightarrow24.3\%\ \text{mAP}{}_{0.5}^{0.95}$).}
\end{figure}

As discussed in section \ref{sec:Approach}, \gnet is directly applicable
to the multi-class setup. We use the exact same parameters and architecture
selected for the persons case. The only change is the replacement
of the score scalar by a per-class score vector in the input and output
(see \S\ref{subsec:Chatty-windows}). We train one multi-class \gnet
model for all 80 COCO categories.

Figure \ref{fig:multiclass-minitest} shows the $\text{mAP}{}_{0.5}^{0.95}$
improvement of \gnet over a per-class tuned GreedyNMS. We obtain
improved results on the bulk of the object classes, and no catastrophic
failure is observed, showing that \gnet is well suited to handle
all kind of object categories. Averaged across classes \gnet obtains
$24.3\%$ $\text{mAP}{}_{0.5}^{0.95}$, compared to $23.5\%$ for
a test-set tuned GreedyNMS. Overall we argue \gnet is a suitable
replacement for GreedyNMS.

Supplementary material includes the detailed per-class table.

\section{\label{sec:Conclusion}Conclusion}

In this work we have opened the door for training detectors that no
longer need a non-maximum suppression (\NMS) post-processing step.
We have argued that \NMS is usually needed as post-processing because
detectors are commonly trained to have robust responses and process
neighbouring detections independently. We have identified two key
ingredients missing in detectors that are necessary to build an \NMS
network: (1) a loss that penalises double detections and (2) joint
processing of detections.

We have introduced the \gnet, the first ``pure'' \NMS network
that is fully capable of performing the \NMS task without having
access to image content nor help from another algorithm. Being a neural
network, it lends itself to being incorporated into detectors and
having access to image features in order to build detectors that can
be trained truly end-to-end. These end-to-end detectors will not require
any post-processing.

The experimental results indicate that, with enough training data,
the proposed \gnet is a suitable replacement for GreedyNMS both for
single- or multi-class setups. The network surpasses GreedyNMS in
particular for occluded cases and provides improved localization.

In its current form the \gnet requires large amounts of training
data and it would benefit from future work on data augmentation or
better initialisation by pre-training on synthetic data. Incorporating
image features could have a big impact, as they have the potential
of informing the network about the number of objects in the image.

We believe the ideas and results discussed in this work point to a
future where the distinction between detector and \NMS will disappear.

\FloatBarrier

\bibliographystyle{ieee}
\bibliography{2017_cvpr_nms_network}

\appendix

\section{\label{sec:Content}Supplementary material}

This supplementary material provides additional details and examples.
Section \ref{sec:Network-details} goes further into detail about
the relation between training and test architecture and about the
detection context layer. Section \ref{sec:postprocessing} illustrates
what raw detections of the detector and the Gnet look like. Section
\ref{sec:qualitative} shows some exemplary detections for GreedyNMS
and Gnet. Section \ref{sec:Coco-persons-test} shows additional COCO
person results. Finally section \ref{sec:Coco-Multi-class} provides
the detailed per-class COCO results.

\clearpage{}

\section{\label{sec:Network-details}Network details}

\paragraph{Training}

\begin{figure}
\subfloat[\label{fig:gnet-high-level-train2}Training architecture.]{\includegraphics[width=1\columnwidth]{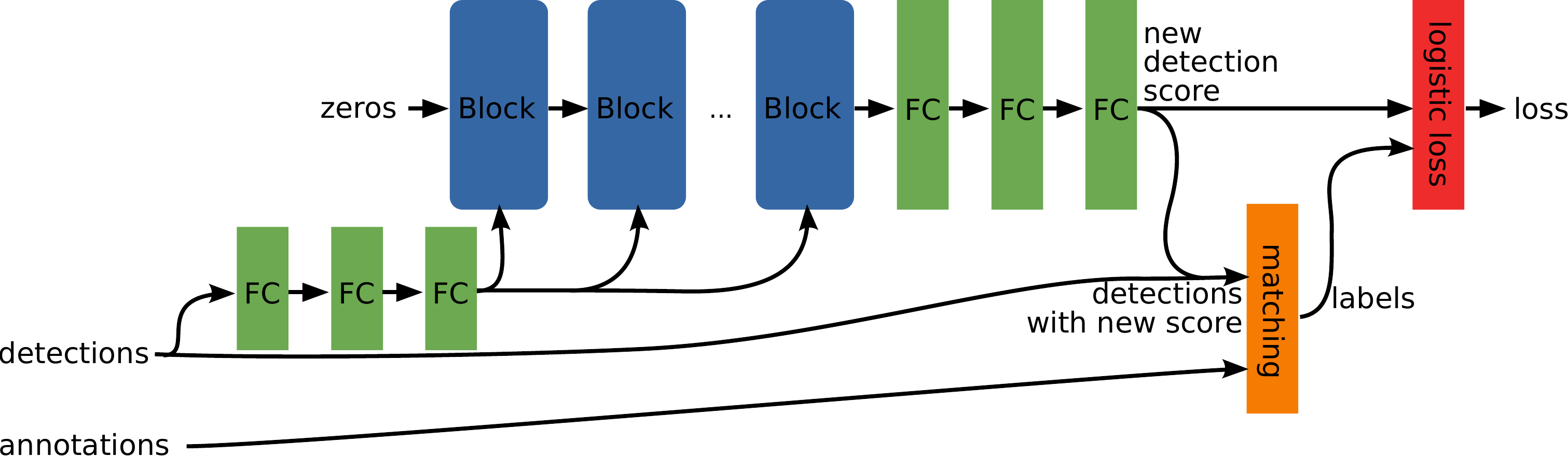}}

\hfill{}\subfloat[\label{fig:gnet-high-level2}Test architecture.]{\includegraphics[width=0.8\columnwidth]{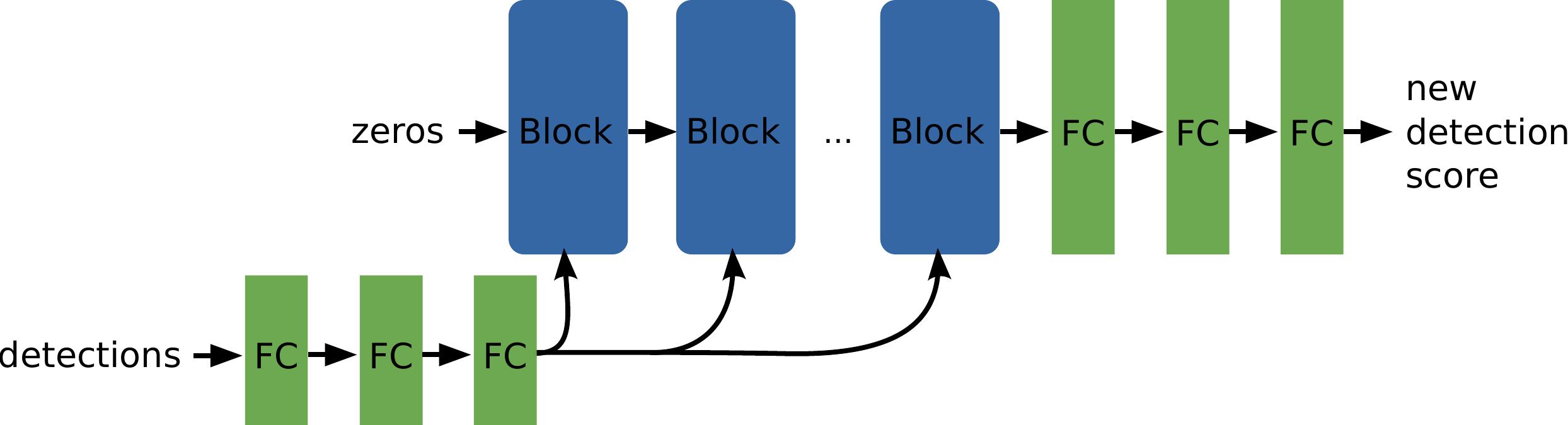}

}\hfill{}

\caption{High level diagram of the Gnet. Blocks as described in section 4.2
of the main paper and in figure 2. FC denotes fully connected layers.
All features in this diagram have 128 dimensions (input vector and
features between the layers/blocks), the output is a scalar.}
\end{figure}

At training time the input of the network consists of detections and
object annotations as illustrated in figure \ref{fig:gnet-high-level-train2}.
The Gnet computes new detections scores for all detections given the
detections only. A detection matching layer takes the detections with
new scores and the object annotations to compute a matching, just
like the benchmark evaluation does. This generates labels: True positives
generate positive labels, false positives negative labels. A logistic
loss layer (SigmoidCrossEntropyLayer in Caffe) takes the new detection
scores and the labels to computes the loss. During backprop the logistic
loss backprops into the Gnet, while the detection matching is assumed
to be fixed and is ignored.\vspace{-1em}

\vspace{1.4em}

\paragraph{Test}

At test time, we remove the detection matching and the loss layer.
The remainder of the network maps detections to new detection scores
and is shown in figure \ref{fig:gnet-high-level2}.

Note that these architectures are identical at training and test time
except for the loss computation. While all state-of-the-art detectors
have an artificial definition of positive and negative detections
at training time and add GreedyNMS at test time, this network is directly
trained for the task and has no post-processing at test time.\vspace{-1em}

\vspace{1.4em}

\paragraph{Pairwise detection context}

\begin{figure}
\includegraphics[width=1\columnwidth]{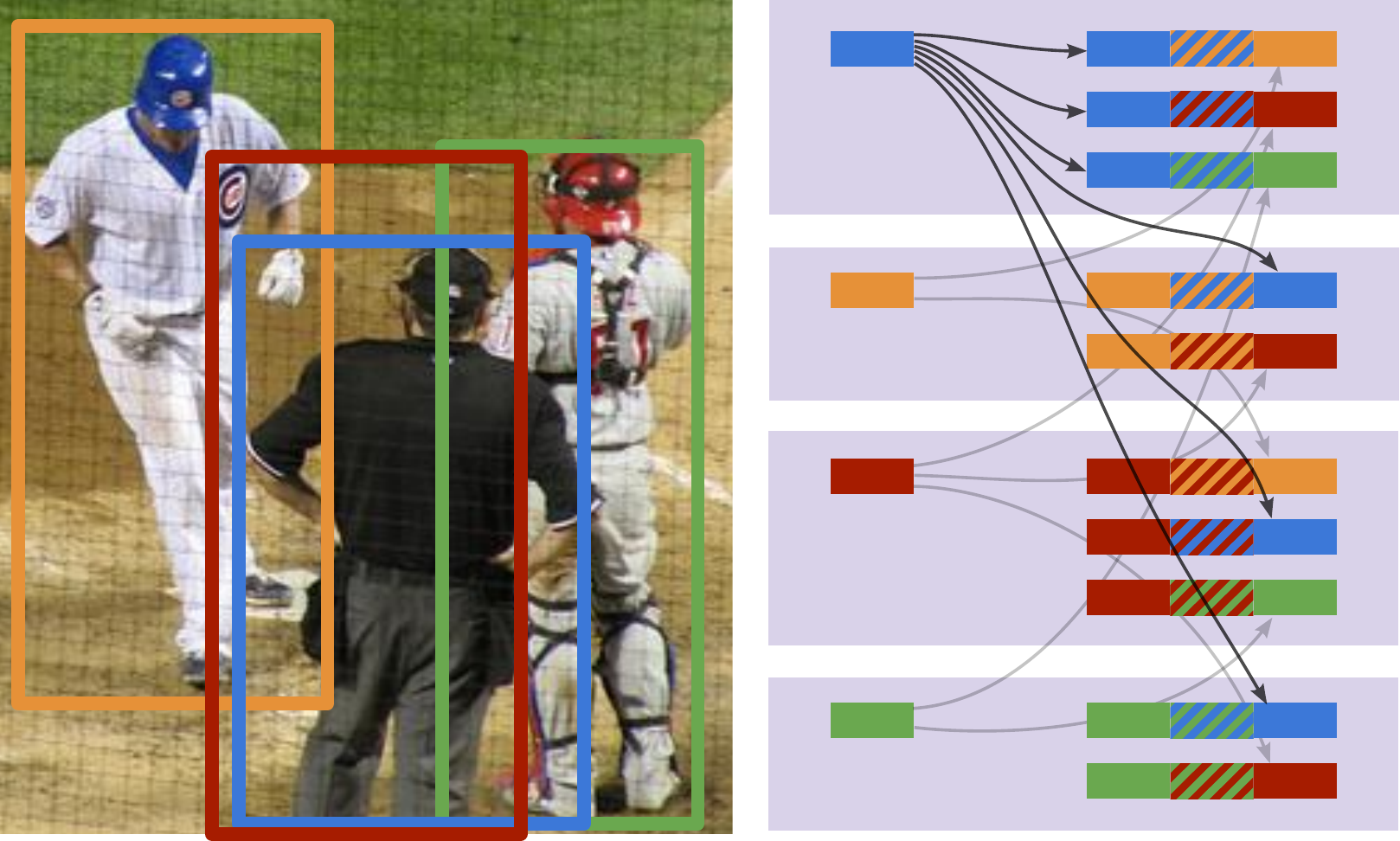}\caption{\label{fig:pairwise-context}Diagram of how detection features are
combined into the pairwise context. Each solid block is the feature
vector of the detection of corresponding colour. The hatched blocks
are the ``detection pair features'' that are defined by the two
detections corresponding to the two colours.}
\end{figure}

Figure \ref{fig:pairwise-context} illustrates the construction of
the pairwise detection context across detections. The feature of the
blue detection is used in the detection context of the blue detection,
but also other detections that overlap with the blue detection. That
means the detection context consists of a variable number of pairs.
The detection context feature for each detection pair is a concatenation
of the detection features (solid boxes) and the detection pair features
(hatched boxes) consisting of properties of the two corresponding
detections as described in the main paper (detection scores, overlap,
relative position, etc.). This combination allows each detection to
access its neighbours feature descriptors and update its own representation
conditioned on its neighbours. Repeating this process can be considered
joint processing of neighbouring detections.

\clearpage{}

\section{\label{sec:postprocessing}Raw detections without post-processing}

\begin{figure}
\hfill{}\subfloat[Raw detections]{\includegraphics[width=0.9\columnwidth]{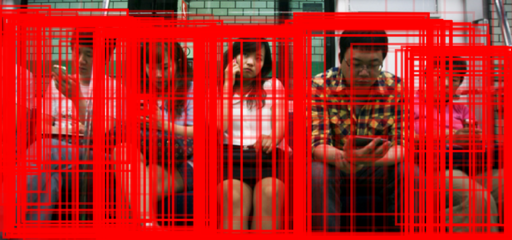}

}\hfill{}

\hfill{}\subfloat[Raw Gnet output]{\includegraphics[width=0.9\columnwidth]{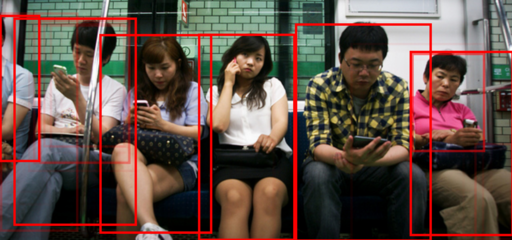}}\hfill{}

\hfill{}\subfloat[Raw detections]{\includegraphics[width=0.9\columnwidth]{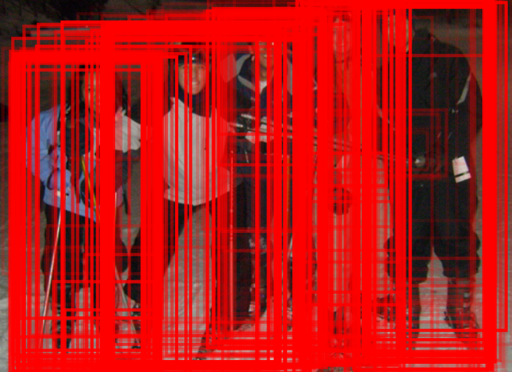}

}\hfill{}

\hfill{}\subfloat[Raw Gnet output]{\includegraphics[width=0.9\columnwidth]{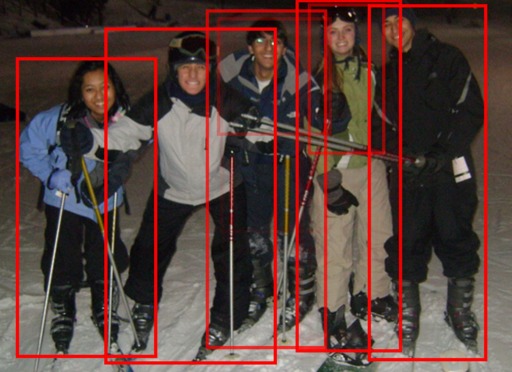}}\hfill{}

\caption{\label{fig:peaky}Raw detections without any post-processing. Detection
opacity is chosen by detection score.}
\end{figure}

To illustrate the fact that the task is non-trivial and that the network
output really needs no post-processing we show raw detections in figure
\ref{fig:peaky}. The opacity of each detection is proportional to
the detection score. Since the detector has soft-max normalised detection
scores and the Gnet has Sigmoid normalised score, we actually choose
the opacity alpha to be equal to the detection score (we don't manually
choose a score transformation that makes unwanted detections perceptually
disappear).

Raw detections are all detections returned by the detector after discarding
very low scoring detections. Note the severe amount of superfluous
detections: people are detected many times and several poorly localized
detections are present.

The raw Gnet detections are not post-processed either. The detector
output and the Gnet output contain the same number of detections,
since the Gnet only rescores detections. Yet the majority of detections
seem to have disappeared, which is due to detections having such a
low score, that they are barely visible. Note how each person has
one clear high scoring detection. Few detections that only detect
the upper body of a person are visible although the person already
was successfully detected, where the Gnet apparently was unsure about
the decision. This is unproblematic as long as these cases are rare
or have a sufficiently low score.

Intuitively, the Gnet gets detections that define a ``blobby'' score
distribution in which similar detections have similar scores into
a ``peaky'' score distribution in which only one detection per object
has a high score and all other detections have a very low score.

\clearpage{}

\section{\label{sec:qualitative}Qualitative results}

\begin{figure}
\hfill{}\subfloat[Gnet]{\includegraphics[width=0.48\columnwidth]{figures/qualitative/qualitative\lyxdot image_0000271402\lyxdot 1}

}\hfill{}\subfloat[GreedyNMS > 0.5]{\includegraphics[width=0.48\columnwidth]{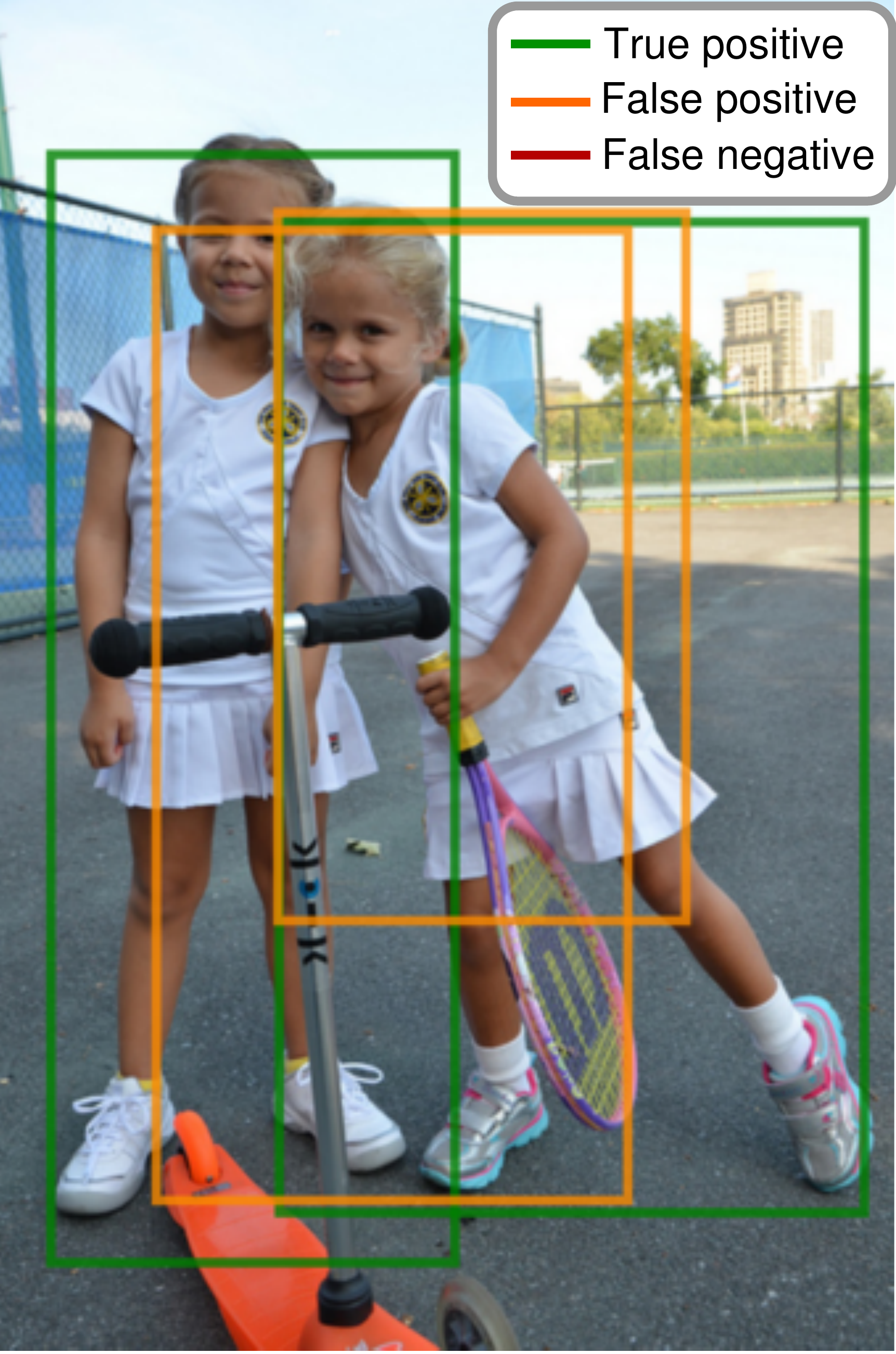}}\hfill{}

\hfill{}\subfloat[Gnet]{\includegraphics[width=0.48\columnwidth]{figures/qualitative/qualitative\lyxdot image_0000072281\lyxdot 1}

}\hfill{}\subfloat[GreedyNMS > 0.5]{\includegraphics[width=0.48\columnwidth]{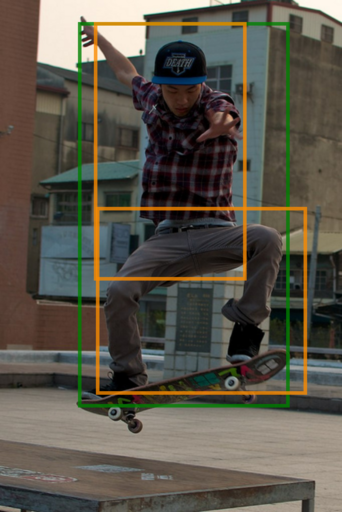}}\hfill{}

\hfill{}\subfloat[Gnet]{\includegraphics[width=0.48\columnwidth]{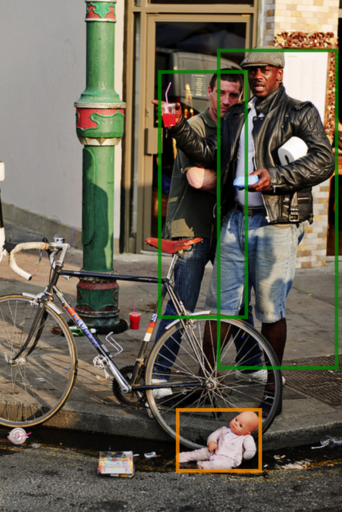}

}\hfill{}\subfloat[GreedyNMS > 0.5]{\includegraphics[width=0.48\columnwidth]{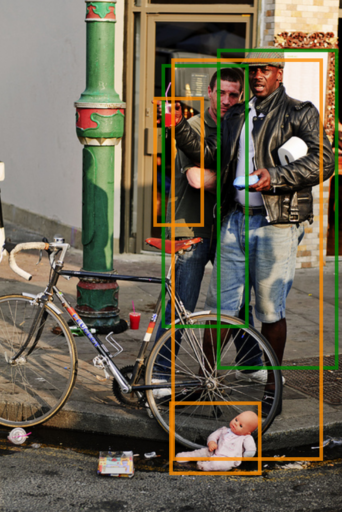}}\hfill{}

\caption{\label{fig:examples-better-suppression}Qualitative results. Both
detectors at the operating point with 60\% recall.}
\end{figure}

\begin{figure*}
\hfill{}\subfloat[Gnet]{\includegraphics[width=0.48\textwidth]{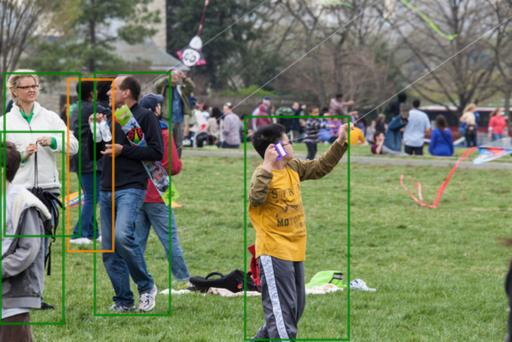}

}\hfill{}\subfloat[GreedyNMS > 0.5]{\includegraphics[width=0.48\textwidth]{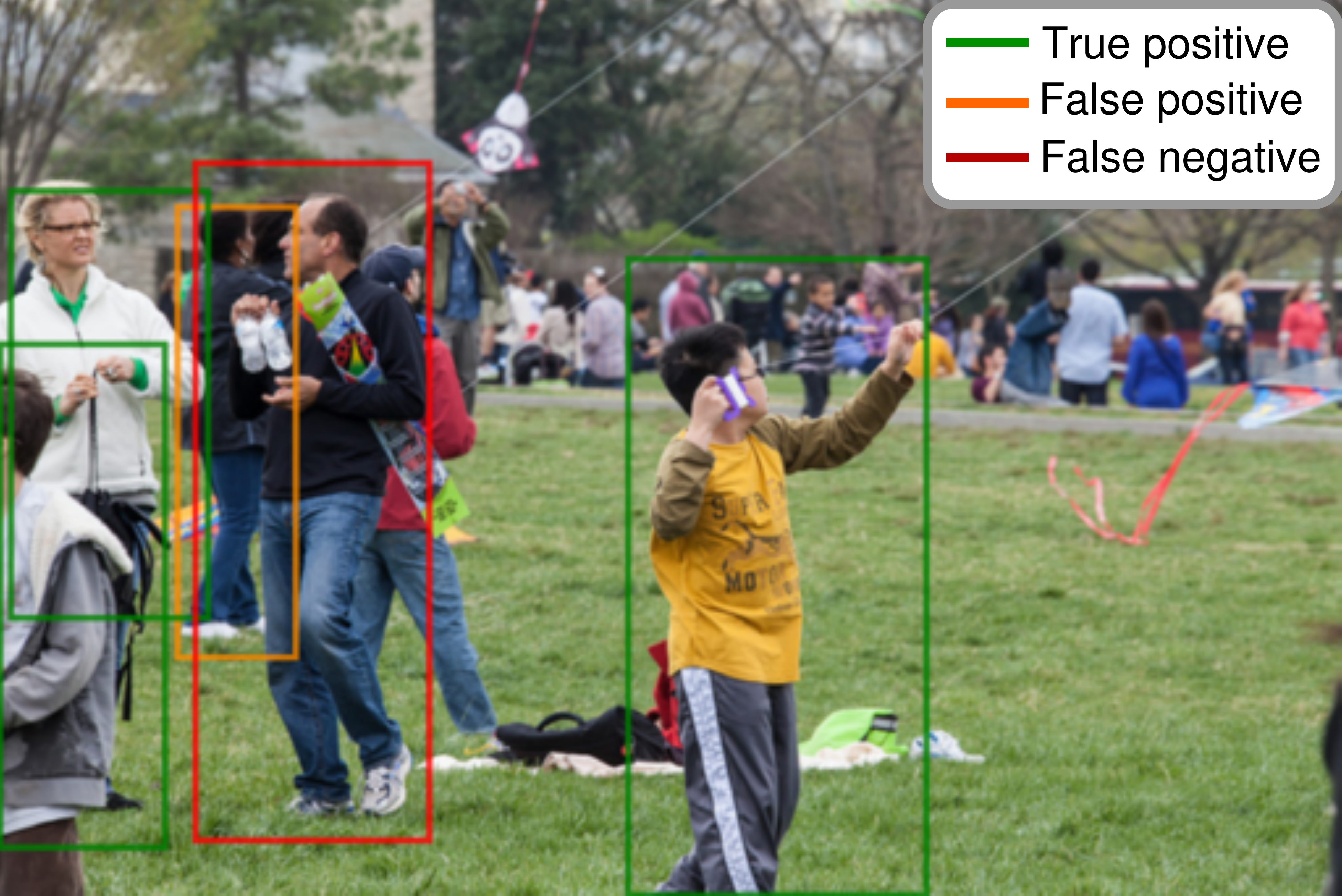}}\hfill{}

\hfill{}\subfloat[Gnet]{\includegraphics[width=0.48\textwidth]{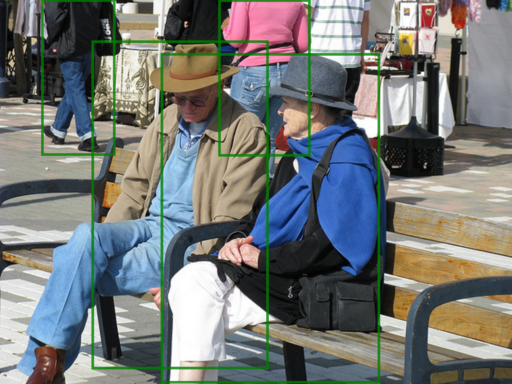}

}\hfill{}\subfloat[GreedyNMS > 0.5]{\includegraphics[width=0.48\textwidth]{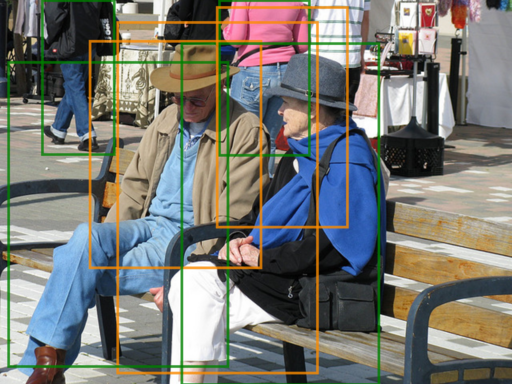}}\hfill{}

\hfill{}\subfloat[Gnet]{\includegraphics[width=0.48\textwidth]{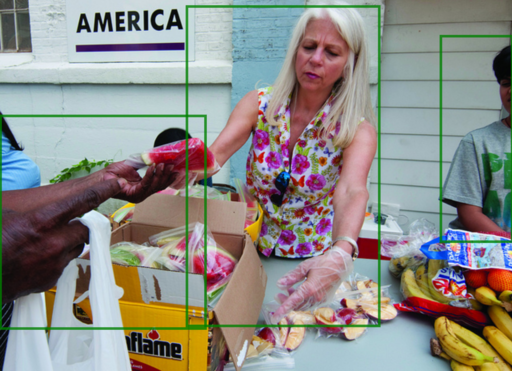}

}\hfill{}\subfloat[GreedyNMS > 0.5]{\includegraphics[width=0.48\textwidth]{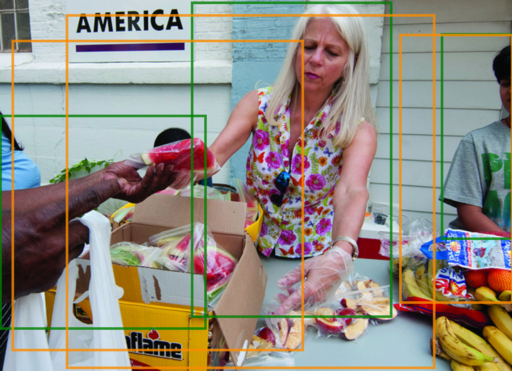}}\hfill{}

\caption{\label{fig:examples1-1}Qualitative results. Both detectors at the
operating point with 60\% recall.}
\end{figure*}

In this section we show exemplary results that compare the Gnet and
GreedyNMS. Both operate on the same set of detections and both are
shown at the same operating point of 60\% recall.

Figures \ref{fig:examples-better-suppression} and \ref{fig:examples1-1}
show that the Gnet is able to suppress maxima that become high scoring
false positives with GreedyNMS. That is the case mostly for detections
that fire on parts of people or too large detections that contain
a person.

Figure \ref{fig:examples1-1} also shows one example of improved recall.
This is possible by increasing the score of an object that had a low
confidence and after applying one specific score threshold is missed.

\clearpage{}

\section{\label{sec:Coco-persons-test}COCO persons mini-test results}

\begin{figure}[h]
\centering{}\includegraphics[width=1\columnwidth]{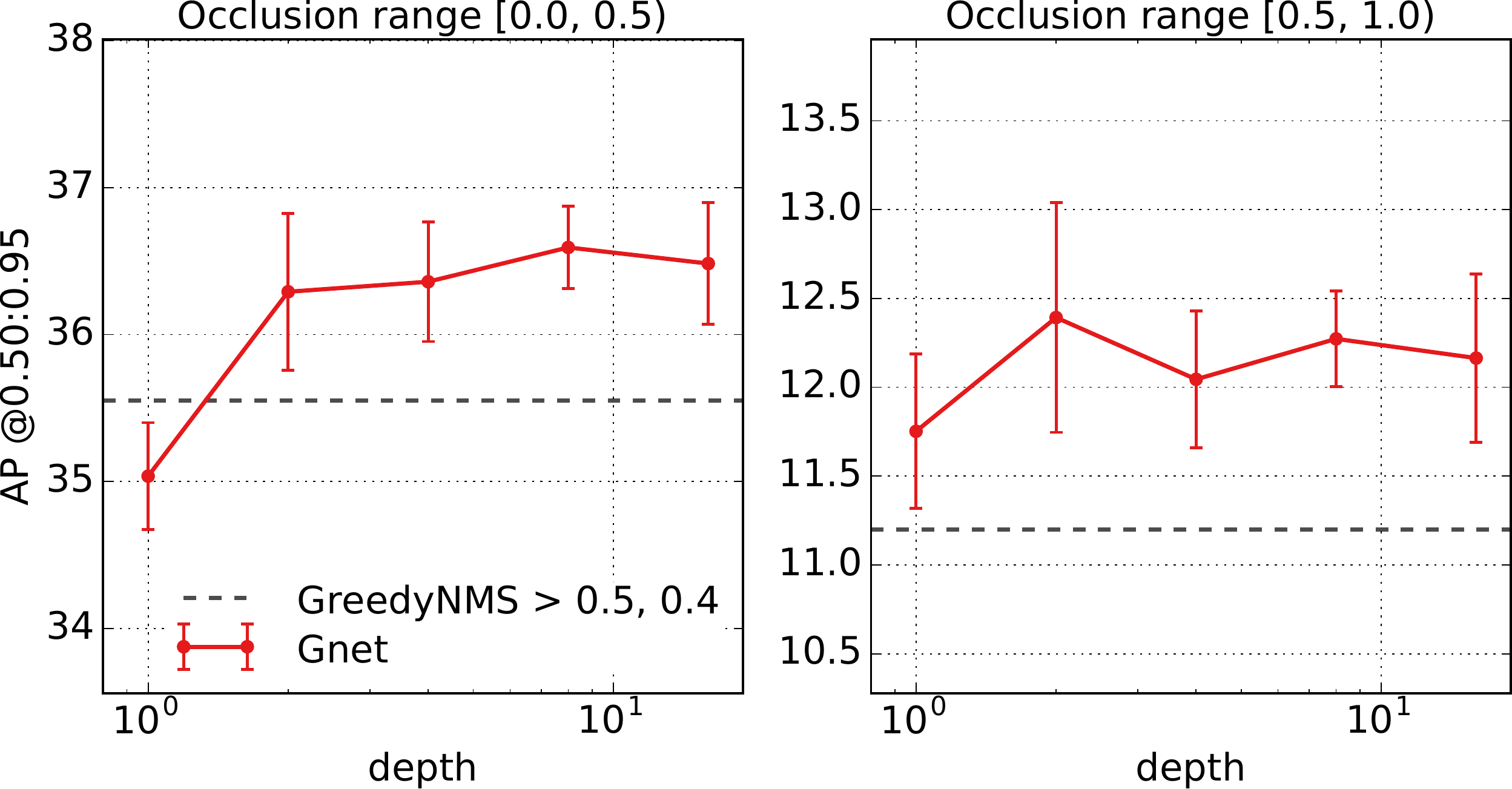}\caption{\label{fig:number-of-blocks-analysis-minitest}$\text{AP}{}_{0.5}^{0.95}$
versus number of Gnet blocks for low and high occlusion respectively
on minitest. Error bars show the standard deviation over six runs.}
\end{figure}

The main paper reports minival results for Gnet models with varying
number of blocks. Figure \ref{fig:number-of-blocks-analysis-minitest}
shows the corresponding results on minitest. We observe the exact
same trend as for minival. One block performs on par to GreedyNMS,
two or more blocks provide a $\sim\negmedspace1\ \text{AP}$ point
gain.

\vspace{3em}

\newpage{}

\section{\label{sec:Coco-Multi-class}COCO multi-class results}

Table \ref{tab:COCO-minitest} provide the detailed per-class improvement
of our multi-class Gnet model over GreedyNMS after tuning its threshold
per-class. Results on COCO minitest set. Averaged across classes Gnet
obtains $24.3\%$ $\text{mAP}{}_{0.5}^{0.95}$, compared to $23.5\%$
for a test-set tuned GreedyNMS.

\begin{table*}
\hfill{}\subfloat{%
\begin{tabular}{l|ccc}
\textbf{\footnotesize{}category} & \textbf{\footnotesize{}GreedyNMS} & \textbf{\footnotesize{}multi-class Gnet} & \textbf{\footnotesize{}improvement}\tabularnewline
\hline 
bed & 30.9 & 34.5 & 3.6\tabularnewline
couch & 24.7 & 28.1 & 3.4\tabularnewline
surfboard & 20.6 & 23.2 & 2.6\tabularnewline
cat & 44.6 & 47.2 & 2.6\tabularnewline
dog & 39.5 & 41.8 & 2.4\tabularnewline
truck & 21.0 & 23.4 & 2.4\tabularnewline
sandwich & 21.6 & 23.8 & 2.1\tabularnewline
car & 21.3 & 23.3 & 2.0\tabularnewline
hot dog & 17.4 & 19.4 & 2.0\tabularnewline
toilet & 42.8 & 44.6 & 1.8\tabularnewline
cow & 29.4 & 31.2 & 1.8\tabularnewline
oven & 22.6 & 24.3 & 1.8\tabularnewline
fork & 10.4 & 12.0 & 1.6\tabularnewline
keyboard & 29.2 & 30.7 & 1.6\tabularnewline
teddy bear & 28.6 & 30.1 & 1.5\tabularnewline
donut & 29.2 & 30.7 & 1.5\tabularnewline
sheep & 29.6 & 31.0 & 1.4\tabularnewline
umbrella & 17.6 & 18.9 & 1.4\tabularnewline
train & 45.0 & 46.3 & 1.3\tabularnewline
bicycle & 18.6 & 19.8 & 1.3\tabularnewline
frisbee & 30.8 & 32.1 & 1.2\tabularnewline
remote & 9.6 & 10.8 & 1.2\tabularnewline
kite & 15.8 & 16.9 & 1.1\tabularnewline
bottle & 15.3 & 16.4 & 1.1\tabularnewline
scissors & 11.2 & 12.3 & 1.1\tabularnewline
skis & 9.4 & 10.5 & 1.1\tabularnewline
cake & 20.5 & 21.6 & 1.1\tabularnewline
bench & 12.8 & 13.9 & 1.1\tabularnewline
pizza & 39.5 & 40.5 & 1.0\tabularnewline
cup & 20.9 & 21.9 & 1.0\tabularnewline
dining table & 23.2 & 24.2 & 1.0\tabularnewline
suitcase & 16.4 & 17.4 & 1.0\tabularnewline
backpack & 6.1 & 7.0 & 1.0\tabularnewline
snowboard & 16.5 & 17.4 & 1.0\tabularnewline
skateboard & 25.9 & 26.8 & 0.9\tabularnewline
motorcycle & 28.3 & 29.2 & 0.9\tabularnewline
toothbrush & 3.8 & 4.6 & 0.8\tabularnewline
bus & 45.9 & 46.7 & 0.8\tabularnewline
refrigerator & 29.2 & 30.0 & 0.8\tabularnewline
horse & 38.1 & 38.9 & 0.8\tabularnewline
\end{tabular}}\hfill{}\subfloat{{\footnotesize{}}%
\begin{tabular}{l|ccc}
\textbf{\footnotesize{}category} & \textbf{\footnotesize{}GreedyNMS} & \textbf{\footnotesize{}multi-class Gnet} & \textbf{\footnotesize{}improvement}\tabularnewline
\hline 
cell phone & 16.1 & 16.9 & 0.8\tabularnewline
broccoli & 16.6 & 17.4 & 0.8\tabularnewline
chair & 13.1 & 13.9 & 0.8\tabularnewline
knife & 4.7 & 5.4 & 0.7\tabularnewline
clock & 31.3 & 32.1 & 0.7\tabularnewline
boat & 13.7 & 14.4 & 0.7\tabularnewline
wine glass & 20.2 & 20.9 & 0.7\tabularnewline
tie & 14.0 & 14.7 & 0.7\tabularnewline
banana & 12.3 & 12.9 & 0.6\tabularnewline
book & 3.5 & 4.2 & 0.6\tabularnewline
handbag & 3.6 & 4.2 & 0.6\tabularnewline
spoon & 4.4 & 5.0 & 0.6\tabularnewline
zebra & 52.8 & 53.3 & 0.6\tabularnewline
vase & 19.8 & 20.3 & 0.5\tabularnewline
bird & 16.8 & 17.3 & 0.5\tabularnewline
traffic light & 11.0 & 11.4 & 0.4\tabularnewline
carrot & 10.5 & 10.9 & 0.4\tabularnewline
elephant & 49.8 & 50.2 & 0.3\tabularnewline
hair drier & 1.7 & 2.1 & 0.3\tabularnewline
bowl & 24.0 & 24.3 & 0.3\tabularnewline
laptop & 38.9 & 39.2 & 0.3\tabularnewline
sink & 20.8 & 21.1 & 0.3\tabularnewline
orange & 17.5 & 17.8 & 0.2\tabularnewline
tv & 38.4 & 38.6 & 0.2\tabularnewline
baseball glove & 16.1 & 16.3 & 0.1\tabularnewline
stop sign & 47.0 & 47.2 & 0.1\tabularnewline
sports ball & 12.3 & 12.4 & 0.1\tabularnewline
airplane & 37.6 & 37.6 & 0.0\tabularnewline
potted plant & 13.9 & 13.8 & -0.0\tabularnewline
baseball bat & 13.2 & 13.1 & -0.1\tabularnewline
parking meter & 19.4 & 19.3 & -0.1\tabularnewline
tennis racket & 29.1 & 28.9 & -0.2\tabularnewline
toaster & 12.2 & 11.8 & -0.3\tabularnewline
microwave & 36.0 & 35.6 & -0.3\tabularnewline
person & 35.5 & 35.1 & -0.4\tabularnewline
apple & 12.5 & 11.9 & -0.6\tabularnewline
mouse & 24.5 & 23.8 & -0.6\tabularnewline
bear & 54.2 & 52.5 & -1.7\tabularnewline
fire hydrant & 44.6 & 42.8 & -1.8\tabularnewline
giraffe & 55.6 & 53.2 & -2.4\tabularnewline
\end{tabular}}\hfill{}

\caption{\label{tab:COCO-minitest}$\text{AP}{}_{0.5}^{0.95}$ per class on
the COCO minitest set, sorted by improvement over GreedyNMS. GreedyNMS
threshold selected optimally per-class on the minitest set.}

\end{table*}

\end{document}